%% file: main.tex
\documentclass[10pt,twocolumn,letterpaper]{article}

\usepackage[pagenumbers]{wacv} 


\usepackage{graphicx}
\usepackage{amsmath}
\usepackage{amssymb}
\usepackage{booktabs}
\usepackage{enumitem}

\usepackage{subcaption}

\definecolor{wacvblue}{rgb}{0.21,0.49,0.74}
\usepackage[pagebackref,breaklinks,colorlinks,allcolors=wacvblue]{hyperref}

\def\wacvPaperID{1539} 
\def\confName{WACV}
\def\confYear{2026}

\title{Divisive Decisions: Improving Salience-Based Training for Generalization in Binary Classification Tasks}

\author{Jacob Piland\\
Department of Computer Science and Engineering, University of Notre Dame \\
Notre Dame, IN 46556 \\
{\tt\small jpiland@nd.edu}
\and
Christopher Sweet\\
Center for Research Computing, University of Notre Dame \\
Notre Dame, IN 46556 \\
{\tt\small csweet1@nd.edu}
\and
Adam Czajka\\
Department of Computer Science and Engineering, University of Notre Dame \\
Notre Dame, IN 46556 \\
{\tt\small aczajka@nd.edu}
}

\begin{document}
\maketitle

\begin{abstract}
   Existing saliency-guided training approaches improve model generalization by incorporating a loss term that compares the model's class activation map (CAM) for a sample's true-class ({\it i.e.}, correct-label class) against a human reference saliency map. However, prior work has ignored the false-class CAM(s), that is the model's saliency obtained for incorrect-label class. We hypothesize that in binary tasks the true and false CAMs should diverge on the important classification features identified by humans (and reflected in human saliency maps). We use this hypothesis to motivate three new saliency-guided training methods incorporating both true- and false-class model's CAM into the training strategy and a novel post-hoc tool for identifying important features. We evaluate all introduced methods on several diverse binary close-set and open-set classification tasks, including synthetic face detection, biometric presentation attack detection, and classification of anomalies in chest X-ray scans, and find that the proposed methods improve generalization capabilities of deep learning models over traditional (true-class CAM only) saliency-guided training approaches. We offer source codes and model weights\footnote{GitHub repository link removed to preserve anonymity} to support reproducible research.
\end{abstract}


\section{Introduction}

\subsection{Background and Motivation} Deep neural networks have demonstrated impressive performance across various computer vision tasks, but their opaque decision-making process remains a significant limitation. 
Salience-based explainability methods, including class activation maps (CAMs) \cite{zhou2016initialCAM}, have been widely adopted to address this issue by visualizing the image regions most influential to a model’s prediction. 
While initially developed for post-hoc interpretability, CAMs have also been incorporated into the saliency-guided training paradigms, where models are rewarded for aligning their attention with human- or auxiliary model-provided annotations. 
One example implementation of such training paradigm is CYBORG method \cite{boyd2023cyborg}, which improves generalization by penalizing discrepancies between the CAM of the true-class and a human salience map.

However, prior work has shown that models can be trained to produce visually persuasive but misleading CAMs for methods utilizing only the true ({\it i.e.}, sample-specific correct-label) class without harming classification accuracy, known as passive fooling \cite{heo2019fooling}. 
This suggests that supervising only the true-class CAM may be insufficient to ensure meaningful model attention.

\subsection{Proposed Solution}

In this work, we revisit the CAMs of both the true and false ({\it i.e.}, sample-specific incorrect-label) classes during training, using what we refer to as the ``teacher'' setup, in which class labels are known and used to generate both CAMs. 
Rather than supervising CAMs in isolation, we propose loss terms that enforce a contrast between the true- and false-class CAMs, either directly or indirectly through human annotations. {\bf We introduce three training variants and one novel visualization approach} that build on this intuition: 

\begin{enumerate}[label=(\alph*)]
\item the first training method {\bf supervises the CAM difference to match human annotations}; we further present this ``Difference Salience'' as a novel CAM that reveals new and plausible features from the contrast of the two classes;

\item the second training method, called in this paper ``Per-class Salience,'' {\bf independently supervises the true and false CAMs to match the human map and its inverse}, respectively;

\item the third proposed method, called ``Contrast Salience,'' {\bf supervises the true-class CAM to match human annotations while encouraging the false-class CAM to diverge from it} by matching an inverted version of the true CAM. 
\end{enumerate}

\subsection{Evaluation Domains}

All three proposed methods aim to induce more discriminative internal representations and improve generalization. We evaluate them in three contexts and domains: 

\begin{itemize}
\item {\bf in-set} chest X-ray anomaly detection, to serve as a baseline domain, in which generalization capabilities of the classifier are not crucial,

\item {\bf } {\bf out-of-set} synthetic face detection, where prior saliency-guided training methods have shown strong out-of-distribution classification accuracy gains, and 

\item {\bf out-of-set} iris presentation attack detection (PAD), which has also been used in previous works to evaluate saliency-guided training. 

\end{itemize} 

\subsection{Research Questions}

We find that while traditional saliency-guided training methods already improve generalization, the addition of contrastive CAM supervision leads to further benefits in challenging generalization settings. To structure our investigations related to concrete benefits coming from the proposed methods, we define the following \textbf{research questions}, around which our experiments are built:

\begin{description}[leftmargin=1cm]
\item[{\bf RQ1:}] Does Difference Salience reveal new and plausible features in models trained to obfuscate their true-class CAMs with passive fooling?
\item[{\bf RQ2:}] In binary classification, does supervising the CAM difference using human annotations improve model generalization beyond traditional saliency-guided training?
\item[{\bf RQ3:}] Does directly supervising both true and false CAMs (per-class salience) using complementary annotations (human-sourced: direct and inverted) improve model behavior?
\item[{\bf RQ4:}] Can contrastive supervision using human-guided true-class CAM to define a target for false-class CAM yield additional gains in classification generalization?
\end{description}

\subsection{Summary of Contributions} 

We propose a novel visualization and saliency-guided training target called Difference Salience and qualitatively demonstrate it's value.

We further propose and evaluate three progressively stronger saliency-guided training methods based on the use of both class CAMs in a binary classification set-up, and applied to both in-set and out-of-set classification problems. First, we modify the loss function to request that the difference between unnormalized CAMs, rather than the true-class CAM alone, match human-sourced salient features (obtained via image annotations or eye tracking). Second, we jointly supervise both CAMs: the true-class CAM is aligned with human saliency, and the false-class CAM is aligned with the inverted human saliency heatmap. 
Third, we supervise the true-class CAM with human saliency, and require the false-class CAM to match the inverted true-class CAM, allowing the model to maintain a strong contrast in class CAMs even when the true-class CAM differs from the human annotations.

Finally, we offer the source codes, all model weights and training configurations (splits, seeds, etc.) along with the paper\footnote{GitHub repository link removed to preserve anonymity} to support the reproducible research.

\section{Past Works and This Study}

Since their introduction in 2016 \cite{Zhou_CVPR_2016}, class activation maps (CAMs) have become a widely used tool for visualizing model decision-making. Numerous CAM variants have been proposed \cite{gradcam,gradcam++,scorecam,ablationcam,xgradcam,eigencam,hirescam}, many of which have been applied both post hoc and during training to improve interpretability and classification generalization. Salience-based training methods, such as CYBORG \cite{boyd2023cyborg}, incorporate human annotations to encourage alignment between model attention, represented by one of the CAMs above, and human-identified salient regions, leading to improved generalization in multiple domain settings \cite{boyd2022humanImproveGeneralization}.

However, recent work has shown that salience can be manipulated through  adversarial methods during training that preserve accuracy while misleading interpretation, a phenomenon known as {\it passive fooling} \cite{heo2019fooling}. This has motivated efforts to improve the reliability of saliency-guided training, either through model design or training objectives.

\textbf{This work differs from previous works} in saliency-guided training by revisiting the role of false-class CAMs, which have been largely ignored in past work. We introduce novel training objectives that explicitly contrast the true- and false-class CAMs using human supervision. Unlike earlier methods that supervise the true-class CAM in isolation, our proposed Difference Salience, Per-class Salience, and Contrast Salience methods aim to improve generalization by promoting discriminative internal representations through {\bf CAM divergence}. Additionally, we present Difference Salience as a novel CAM visualization that captures decision-critical regions even in passively-fooled models.

\section{Difference Salience}

\subsection{Calculation Method}

Traditional CAM relies only on the activations of the true class. We calculate the Difference Salience $d_k^{\text{norm}}$ for the $k$-th sample by estimating both the true- and false-class CAMs for that sample and subtracting them before normalization:
\vskip-5mm
\begin{equation}
\begin{split}
    d_k^{\text{norm}} = \text{norm}_{[0,1]}(t_k - f_k)     
\end{split}
\label{eqn:dk}
\end{equation}

\noindent
where $t_k$ is the unnormalized true-class CAM ({\it i.e.}, the CAM for the $k$-th sample's correct label), $f_k$ is the unnormalized false-class CAM ({\it i.e.}, the CAM for the sample's incorrect label), and $\text{norm}_{[0,1]}(x) = (x - \min(x))/(\max(x) - \min(x))$ remaps $x$ to a unite interval. Each class-dependent CAM is composed of the pixel-wise sum of the feature weights for that class multiplied by the input to the final classification (linear) layer of neurons in the classification model. 

As these are the values that are used to calculate the logits for the classifier, it is their difference that decides an input's classification label.

\subsection{Visualization and Use of Difference Salience}

Passively fooled models are those, which are trained to obfuscate the activations in their true-class CAMs.
As it is the difference in the logits that determine a model's decision, we hypothesize it is the difference in class activations (which directly contribute to logit calculation) for a region, which will more correctly highlight which image regions contributed to a model's decision.
The Difference Salience, $d_k^{\text{norm}}$, is a CAM that can be used as any other CAM in saliency-guided training and can be visualized as any other CAM to highlight decision-critical regions of an image (see Fig. \ref{fig:cam_grid} for example visualizations made for samples representing three domains evaluated in this paper).

\section{Saliency-Guided Training Methods}
\subsection{Baseline}

A traditional saliency-guided training loss function using human annotations consists of a classification element and human perception element:

\begin{equation}
\label{eqn:CYBORGloss}
\begin{split}
\mathcal{L}_{\text{Baseline}}=
\text{$-\alpha$ $\underbrace{\log p^{(m)}(y_k \in C)}_{\text{classification component}}$}\\
\text{$+\beta$ $\underbrace{\text{MSE}(h_k^{\text{norm}},t_k^{\text{norm}})}_{\text{human perception component}}$}
\end{split}
\end{equation}

\noindent
where $y_k$ is the correct class label for the $k$-th sample, $C$ is the set of class labels, $h_k^{\text{norm}}$ is the human saliency map remapped to a unite interval, and $t_k^{\text{norm}}$ is the normalized true-class CAM. The weighting parameters are $\alpha$ for the cross-entropy-based loss component and $\beta$ for the human saliency-based loss component. Our proposed methods use a similar structure, but replace or add to the human perception component one based on our CAM difference observation.

\subsection{Novel Method 1 (Difference Salience)}

Our first method replaces the true-class CAM $t_k^{\text{norm}}$ in \cref{eqn:CYBORGloss} with CAM difference $d_k^{\text{norm}}$ from \cref{eqn:dk}:

\begin{equation}
\label{eqn:nm1}
\begin{split}
\mathcal{L}_{\text{Difference Salience}}=
-\alpha \log p^{(m)}(y_k \in C)\\+\beta\text{MSE}(h_k^{\text{norm}},d_k^{\text{norm}})
\end{split}
\end{equation}


\noindent
This forces the model's true- and false-class activations to diverge most strongly where the expert human annotations indicate important features.

\subsection{Novel Method 2 (Per-class Salience)}

Our second method adds to the human perception component from \cref{eqn:CYBORGloss} another component $\text{MSE}(1-h_k^{\text{norm}},f_k^{\text{norm}})$ to additionally and independently supervise the false-class CAM to the inverse of heatmap representing human saliency:

\begin{equation}
\label{eqn:nm2}
\begin{split}
\mathcal{L}_{\text{Per-class Salience}}=
-\alpha \log p^{(m)}(y_k \in C)\\+\beta\text{MSE}(h_k^{\text{norm}}, t_k^{\text{norm}}) + \gamma\text{MSE}(1-h_k^{\text{norm}},f_k^{\text{norm}})
\end{split}
\end{equation}


\noindent
where $\gamma$ serves as the weighting parameter for the third component and $f_k^{\text{norm}}$ is the normalized false-class CAM.

We hypothesize that because using the human salience to guide the model to important features for the true-class CAM improves performance, jointly requesting the model match the inverse of the annotations for the false-class CAM will strengthen the difference in class activations and improve the model's generalization capabilities further.

\subsection{Novel Method 3 (Contrast Salience)}
Requiring both true-class CAM to match the human annotations and false-class CAM to match the inverse, as proposed in the second method above, places an extra importance on the human annotations. 
The model may become less able to diverge from the human annotations where needed. Thus, the third novel method emphasizes the difference in class activations while only guiding the true-class CAM with human salience:

\begin{equation}
\label{eqn:nm3}
\begin{split}
\mathcal{L}_{\text{Contrast Salience}}=
-\alpha \log p^{(m)}(y_k \in C)\\+\beta\text{MSE}(h_k^{\text{norm}}, t_k^{\text{norm}}) + \gamma\text{MSE}(1-t_k^{\text{norm}},f_k^{\text{norm}})
\end{split}
\end{equation}


This allows the model more leeway for fine-tuning the activation weights, while still rewarding the divergence of the false-class CAM.

\section{Experimental Design}
\label{sec:exp_design}

\subsection{Experiments Addressing Research Questions}

We conduct four experiments: 

\begin{enumerate}[label=(\alph*)]
    \item training passively-fooled models and extracting sample salience from each class (including Difference Salience) for comparison,
    \item supervising CAM difference in binary classification (addressing {\bf RQ1}), 
    \item directly supervising both true- and false-class CAMs with human annotations and their inverse (addressing {\bf RQ2}), and
    \item using contrastive supervision, guiding the true CAM with human annotations and requesting the false-class CAM match the inverse of the true-class CAM (addressing {\bf RQ3}).
\end{enumerate}

We use an established baseline saliency-guided training method \cite{boyd2023cyborg} for comparison. 

\subsection{Training Scenarios and Performance Metrics}

For experiment (a) we train one instance of each model for each domain using passive fooling to direct the CAMs toward the edges of the model. We use \cref{eqn:CYBORGloss} as the loss function with a false human salience annotating the image edges.
The remaining model trainings have the same experimental format.
We train ten instances of each model for each domain.
We compare the performance using the Area Under the Receiver Operating Characteristic Curve (AUROC).
We use AUROC as this metric was used in the traditional saliency-guided training upon which we directly build and with which we must compare.

\subsection{Experiment Parameters}
\label{ssec:exp_params}

All models are instantiated from the DenseNet-121 architecture \cite{densenet121_2017}, which has been pre-trained on ImageNet dataset.
All models are trained for 50 epochs using Stochastic Gradient Descent with a learning rate of 0.002 and a different random seed. The weighting components for all loss functions are equal. For models with two components $\alpha=\beta=0.5$ and for models with three components $\alpha=\beta=\gamma=0.3$.

\subsection{Datasets}
\label{ssec:datasets}

We use the samples from existing datasets in their respective tasks according to Table \ref{tab:datasets}.

Chest X-ray images may either be entirely normal or contain one or more of the following: Atelectasis, Cardiomegaly, Edema, Lung Opacity, Pleural Effusion, Pneumonia, and Support Devices. 

Iris PAD models are trained with a leave-one-out method where all but one attack type is used in the training set and the remaining one is used for the testing set. Alongside the real iris category \cite{real_iris1,real_artificial_textured_print_24,diseased38,real22,real_textured20,real_textured46,real43,real_textured45,boyd2022humanImproveGeneralization}, the PAD types are: artificial (e.g., glass prosthetics)  \cite{real_artificial_textured_print_24,boyd2022humanImproveGeneralization}, Textured Contacts \cite{real_artificial_textured_print_24,real_textured20,real_textured46,real_textured45,boyd2022humanImproveGeneralization}, Post-Mortem \cite{post40}, Printouts \cite{real_print12,real_artificial_textured_print_24,print21}, Printouts with contacts\cite{real_artificial_textured_print_24}, Synthetic \cite{synth44}, and Diseased \cite{diseased38}.

The synthetic face detection set is the established dataset from \cite{boyd2023cyborg} which provides our baseline model. We use this dataset with no change to the training or testing partitions in order to make the most valid comparison. Models are trained with limited overlap between the training and testing sets. The training set consists of real samples from the Face Recognition Grand Challenge (FRGC) dataset \cite{FRGC} and synthetic samples from the Synthesis of Realistic Face Images (SREFI) benchmark  \cite{SREFI} and synthesized by StyleGAN2 \cite{stylegan2020}. The test sets include: real images from CelebA-HQ \cite{celeba_hq2017} and Flicker-Faces-HQ (FFHQ) \cite{FFHQ2018} and synthetic ones generated using ProGAN \cite{progan2019}, StarGANv2 \cite{stargan2018}, StyleGAN \cite{stylegan2019}, StyleGAN2 \cite{stylegan2020}, StyleGAN3 \cite{stylegan2021}, and StyleGAN2-ADA \cite{stylegan_ada2020}. 

\begin{table*}[ht]
    \centering
    \caption{A summary of the datasets, number of samples and tasks considered in this paper. }
    \begin{tabular}{c c c c c}
        \toprule
        \textbf{Dataset} & \textbf{Task} & \textbf{Training} & \textbf{Testing} & \textbf{Source(s)} \\
        \midrule
        Chest & In set & 667 (normal), & 54,836 (normal), & \cite{mimicxr} \\
        X-rays &       & 1,161 (abnormal) & 110,469 (abnormal) & \\
        \midrule
        Iris PAD & Out of set  & 1,351 (real),  & 11,551 (real),  & \cite{real_iris1,real_artificial_textured_print_24,real_print12,real_diseased,real22,real_textured20,real_textured46,real43,real_textured45,post40,print21,synth44,diseased38} \\
         &   & 3937 (spoof) & 11510 (spoof) & \\
         \midrule
        Synthetic & Out of set  & 919 (real),  & 100,000 (real), & \cite{FRGC,SREFI,stylegan2020,celeba_hq2017,FFHQ2018,progan2019,stargan2018,stylegan2019,stylegan2021,stylegan_ada2020} \\
        face detection &  & 902 (spoof) &  600,000 (spoof) & \\
        \bottomrule
    \end{tabular}
    
    \label{tab:datasets}
\end{table*}

\section{Results}

\begin{figure*}[htbp]
    \centering
    \begin{subfigure}{0.48\textwidth}
        \includegraphics[width=\linewidth]{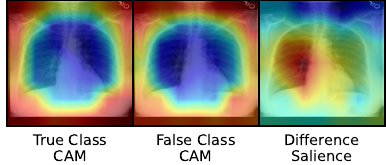}
        \caption{Chest X-rays normal sample}
    \end{subfigure} \hfill
    \begin{subfigure}{0.48\textwidth}
        \includegraphics[width=\linewidth]{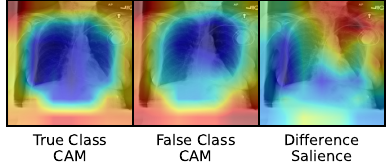}
        \caption{Chest X-rays abnormal sample}
    \end{subfigure} \\
    \vspace{1em}
    \begin{subfigure}{0.48\textwidth}
        \includegraphics[width=\linewidth]{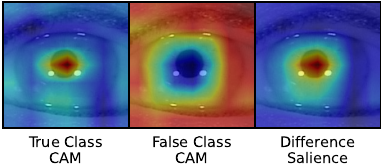}
        \caption{Iris PAD real sample}
    \end{subfigure} \hfill
    \begin{subfigure}{0.48\textwidth}
        \includegraphics[width=\linewidth]{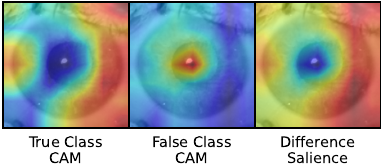}
        \caption{Iris PAD spoof sample}
    \end{subfigure} \\
    \vspace{1em}
    \begin{subfigure}{0.48\textwidth}
        \includegraphics[width=\linewidth]{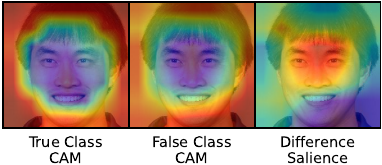}
        \caption{Synthetic face detection real sample}
    \end{subfigure} \hfill
        \begin{subfigure}{0.48\textwidth}
        \includegraphics[width=\linewidth]{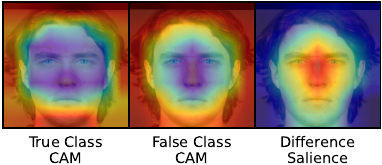}
        \caption{Synthetic face detection spoof sample}
    \end{subfigure}
    \caption{Illustrative examples from three binary class datasets of misleading CAMs produced by models training with passive fooling and how Difference Salience can modify the used features. Each subfigure shows the CAM for the sample's correct label (True-Class CAM), the CAM for the sample's incorrect label (False-Class CAM), and the Difference Salience created by subtracting the unnormalized False-Class CAM from the unnormalized True-Class CAM. Each is image uses a red to blue color scale to indicate regions of higher interest which is separate for each CAM ({\it i.e.}, a red region in the True and False-Class CAMs indicate that it is of higher interest within that CAM only, not that the values are same between the two CAMs).}
    \label{fig:cam_grid}
\end{figure*}

\begin{table}[ht]
    \centering
    \caption{AUROC performance of all methods considered on the three datasets tested. }
    \begin{tabular}{c c c}
        \toprule
        \textbf{Dataset} & \textbf{Method} & \textbf{AUROC} \\
        \midrule
                       & Baseline     & \textbf{0.866±0.005}    \\
        Chest          & Difference Salience      & 0.857±0.005    \\
        X-rays         & Per-class Salience   & 0.856±0.003    \\
                       & Contrast Salience   & 0.855±0.002    \\
        \midrule
                       & Baseline     & 0.786±0.091   \\
        Iris           & Difference Salience      & 0.770±0.097   \\
                       & Per-class Salience   & 0.752±0.103   \\
                       & Contrast Salience   & \textbf{0.790±0.089}   \\
        \midrule
                       & Baseline     & 0.602±0.029    \\
        Synthetic      & Difference Salience      & \textbf{0.651±0.041}    \\
        Face           & Per-class Salience   & \textbf{0.651±0.023}    \\
                       & Contrast Salience   & 0.609±0.029    \\
        
        \bottomrule
    \end{tabular}
    
    \label{tab:auroc}
\end{table}
\subsection{\textbf{Answering RQ1 (Does Difference Salience reveal new and plausible features in models trained to obfuscate their true-class CAMs with passive fooling?)}}

We qualitatively compare illustrative examples in \cref{fig:cam_grid}.
The models used to generate this salience where all trained with passive fooling, {\it i.e.}, trained to create obfuscated CAMs that point toward arbitrary regions (in this case the edges of the image) without impacting model performance.
We see that in every case, the Difference Salience captures some features that distinguish it from the True and False-Class CAMs.

While the Iris PAD model shows some resilience to the passive fooling (the True-Class Salience for the spoof sample is only minorly distorted), most of the models have been successfully fooled and their True-Class Salience indicates the arbitrary edges (False-Class salience need not indicate the edges for successful fooling although many do).

In contrast to the True-Class Salience, the Difference Salience tends to indicate more plausible features away from the edges of the image.
In both chest X-ray samples and both synthetic face samples, the True-Class Salience is either not indicating the target region at all or only capturing the very edge of it.
The Difference Salience clearly captures the actual subject. 
It indicates the actual chest for the normal chest X-ray sample and highlights the round medical device and wires stretching from it in the upper right corner of the abnormal sample.
For faces, it indicates the peri-ocular region instead of the chin and hair.

Iris PAD is not a human-trivial task and it can be difficult to determine which regions are actually important.
We note that the Difference Salience does capture at least some features that the True-Class Salience does not.

Thus, our answer to \textbf{RQ1 is affirmative. Difference Salience reveals new and plausible features even in models trained to obfuscate their CAMs.}

\subsection{\textbf{Answering RQ2 (In binary classification, does supervising the CAM difference using human annotations improve model generalization beyond traditional saliency-guided training?)}}

Quantitatively we see in \cref{tab:auroc} that supervising the CAM difference using human annotations does not improve model performance in the baseline in set task and only improves auroc performance in one of the generalization tasks (synthetic face detection improves  by $8.1\%$ from $0.602$ to $0.651$). This leads us to conclude that the answer to \textbf{RQ2 is task domain specific with the potential to noticeably improve model generalization}.

\subsection{\textbf{Answering RQ3 (Does directly supervising both true and false CAMs (per-class salience) using complementary annotations (human and inverted) improve model behavior?)}}

Similar to the models tested for RQ2, directly supervising both true and false CAMs does not improve model performance in the baseline in set task. Furthermore, this method does not improve iris PAD detection either overall or on any PAD type (See \cref{tab:iris_ablation}). However, it does improve synthetic face detection by $8.1\%$ from $0.602$ to $0.651$. Thus we conclude that the answer to \textbf{RQ3 is task domain specific with the potential to noticeably improve model generalization}.

\subsection{\textbf{Answering RQ4 (Can contrastive supervision using the human-guided true CAM to define a target for the false CAM yield additional gains in generalization?)}}

The contrastive salience method is not necessary for the baseline in set task and does not result in improvement. However, this method improves on AUROC for both generalization tasks. For Iris PAD, our third methods improves AUROC performance for all seven subsets and overall by $0.5\%$. For synthetic face detection, contrastive salience improves AUROC performance for two of six subset and overall AUROC performance by $1.2\%$. Thus we conclude that the answer to \textbf{RQ4 is affirmative, contrastive supervision using the human annotations improves model performance in generalization.}

\begin{table}[ht]
    \centering
    \caption{Ablation study for Iris PAD task. AUROC results are for each model on each data subset. Subsets are named for the attack type left out during training.}
    \begin{tabular}{c c c}
        \toprule
        \textbf{Subset}    & \textbf{Method}       & \textbf{AUROC}        \\
        \midrule
                            & Baseline     & 0.685±0.056    \\
        Artificial          & Difference Salience   & 0.662±0.063    \\
                            & Per-class Salience    & 0.641±0.053    \\
                            & Contrast Salience     & \textbf{0.691±0.060}    \\
        \midrule
                            & Baseline     & 0.706±0.062    \\
        Contacts            & Difference Salience   & 0.682±0.073    \\
         +Print             & Per-class Salience    & 0.649±0.047    \\
                            & Contrast Salience     & \textbf{0.712±0.057}    \\
        \midrule
                            & Baseline     & 0.730±0.048    \\
        Diseased            & Difference Salience   & 0.710±0.054    \\
                            & Per-class Salience    & 0.693±0.046    \\
                            & Contrast Salience     & \textbf{0.735±0.051}    \\
        \midrule
                            & Baseline     & 0.718±0.050    \\
        Post-               & Difference Salience   & 0.698±0.056    \\
        mortem              & Per-class Salience    & 0.675±0.049    \\
                            & Contrast Salience     & \textbf{0.724±0.054}    \\
        \midrule
                            & Baseline     & 0.887±0.025    \\
        Printouts           & Difference Salience   & 0.880±0.025    \\
                            & Per-class Salience    & 0.862±0.028    \\
                            & Contrast Salience     & \textbf{0.889±0.023}    \\
        \midrule
                            & Baseline     & 0.858±0.025    \\
        Synthetics          & Difference Salience   & 0.848±0.028    \\
                            & Per-class Salience    & 0.838±0.024    \\
                            & Contrast Salience     & \textbf{0.861±0.027}    \\
        \midrule
                            & Baseline     & 0.919±0.014    \\
        Textured            & Difference Salience   & 0.912±0.017    \\
         Contacts           & Per-class Salience    & 0.904±0.012    \\
                            & Contrast Salience     & \textbf{0.920±0.016}    \\
        \bottomrule
    \end{tabular}
    \label{tab:iris_ablation}
\end{table}

\begin{table}[]
    \centering
    \caption{Ablation study for synthetic face detection. AUROC results are for each model on each data subset. Each test partition consisted of the same real images and the named synthetic generator samples.}
    \begin{tabular}{c c c}
    \toprule
    \textbf{Subset} & \textbf{Method}           & \textbf{AUROC}        \\
    \textbf{(generator)} & & \\
    \midrule
    StarGAN          & Baseline         & 0.376±0.049    \\
              & Difference Salience       & 0.348±0.058    \\
              & Per-class Salience        & 0.351±0.039    \\
              & Contrast Salience         & \textbf{0.452±0.068}    \\
    \midrule
    ProGAN           & Baseline         & \textbf{0.576±0.024}    \\
               & Difference Salience       & 0.555±0.031    \\
               & Per-class Salience        & \textbf{0.576±0.022}    \\
               & Contrast Salience         & 0.557±0.012    \\
    \midrule
    StyleGAN         & Baseline         & 0.637±0.032    \\
             & Difference Salience       & \textbf{0.710±0.050}    \\
             & Per-class Salience        & 0.704±0.035    \\
             & Contrast Salience         & 0.624±0.026    \\
    \midrule
    StyleGAN2        & Baseline         & 0.713±0.048    \\
            & Difference Salience       & \textbf{0.804±0.058}    \\
            & Per-class Salience        & 0.801±0.031    \\
            & Contrast Salience         & 0.715±0.038    \\
    \midrule
    StyleGAN3        & Baseline         & 0.602±0.053    \\
            & Difference Salience       & \textbf{0.693±0.079}    \\
            & Per-class Salience        & 0.678±0.044    \\
            & Contrast Salience         & 0.599±0.041    \\
    \midrule
    StyleGAN2-ADA    & Baseline         & 0.710±0.046    \\
        & Difference Salience       & \textbf{0.797±0.061}    \\
        & Per-class Salience        & 0.796±0.031    \\
        & Contrast Salience         & 0.707±0.039    \\
    \bottomrule

    \end{tabular}
    \label{tab:face_ablation}
\end{table}

\section{Limitations and Future Work}

While our results demonstrate the potential benefits of incorporating false-class CAMs into saliency-guided training, several limitations should be acknowledged. First, our experiments are restricted to binary classification tasks; extending these techniques to multi-class problems may present both computational and conceptual challenges, particularly in defining contrastive CAM targets when more than one false-class is present.

Second, our use of human salience annotations assumes a reliable correspondence between human visual attention and meaningful classification cues. In domains where this correspondence is weak or ambiguous, performance may degrade. We plan to expand to using human eye tracking and auditory annotations as suitable datasets become available. We also consider AI generated salience as publicly available models improve on their task comprehension.

In further future work, we plan to explore several extensions. One direction is the generalization to multi-class settings, potentially using pairwise CAM contrasts or embedding-based objectives. Another is to investigate the use of learned or model-generated salience proxies in place of human annotations. Finally, integrating these contrastive objectives with adversarial robustness techniques or uncertainty estimation could yield models that are not only more generalizable but also more trustworthy.

\section{Conclusion}

This work revisits saliency-guided training by incorporating supervision over both the true and false-class CAMs in binary classification tasks. Motivated by the hypothesis that meaningful model attention requires not just alignment with important features but also a clear divergence between class-specific salience maps, we propose three novel loss formulations: Difference Salience, Per-class Salience, and Contrast Salience. It further introduces using Difference Salience not only in training but as a post-hoc tool for determining important features that is resistant to passive fooling (a method of training models to produce obfuscated CAMs).

Empirical evaluation across three domains, one in set baseline (chest X-rays) and two generalization tests (iris PAD and synthetic face detection), demonstrates that these methods can improve model generalization beyond traditional salience training. Notably, the Contrast Salience method performs competitively across all domains, achieving the best AUROC scores for iris PAD and improving on the baseline for synthetic face detection. Our results underscore that contrasting CAM behavior, especially with respect to human salience, provides a promising avenue for improving model generalization in decision tasks. Together, these findings support a broader view of saliency-guided supervision: one that not only encourages what a model should attend to, but also discourages what it should not.


{
    \small
    \bibliographystyle{ieeenat_fullname}
    \bibliography{main}
}

\end{document}


%% file: main.bbl
\begin{thebibliography}{37}
\providecommand{\natexlab}[1]{#1}
\providecommand{\url}[1]{\texttt{#1}}
\expandafter\ifx\csname urlstyle\endcsname\relax
  \providecommand{\doi}[1]{doi: #1}\else
  \providecommand{\doi}{doi: \begingroup \urlstyle{rm}\Url}\fi

\bibitem[rea()]{real_iris1}
Chinese academy of sciences institute of automation.
\newblock Accessed: 03-12-2021.

\bibitem[Banerjee et~al.(2017)Banerjee, Bernhard, Scheirer, Bowyer, and Flynn]{SREFI}
Sandipan Banerjee, John~S Bernhard, Walter~J Scheirer, Kevin~W Bowyer, and Patrick~J Flynn.
\newblock {SREFI: Synthesis of realistic example face images}.
\newblock In \emph{IEEE Int. Joint Conf. on Biometrics (IJCB)}, pages 37--45. IEEE, 2017.

\bibitem[Boyd et~al.(2022)Boyd, Bowyer, and Czajka]{boyd2022humanImproveGeneralization}
Aidan Boyd, Kevin~W Bowyer, and Adam Czajka.
\newblock Human-aided saliency maps improve generalization of deep learning.
\newblock In \emph{Proceedings of the IEEE/CVF Winter Conference on Applications of Computer Vision}, pages 2735--2744, 2022.

\bibitem[Boyd et~al.(2023)Boyd, Tinsley, Bowyer, and Czajka]{boyd2023cyborg}
Aidan Boyd, Patrick Tinsley, Kevin~W Bowyer, and Adam Czajka.
\newblock Cyborg: Blending human saliency into the loss improves deep learning-based synthetic face detection.
\newblock In \emph{Proceedings of the IEEE/CVF Winter Conference on Applications of Computer Vision}, pages 6108--6117, 2023.

\bibitem[Chattopadhay et~al.(2018)Chattopadhay, Sarkar, Howlader, and Balasubramanian]{gradcam++}
Aditya Chattopadhay, Anirban Sarkar, Prantik Howlader, and Vineeth~N Balasubramanian.
\newblock Grad-cam++: Generalized gradient-based visual explanations for deep convolutional networks.
\newblock In \emph{2018 IEEE winter conference on applications of computer vision (WACV)}, pages 839--847. IEEE, 2018.

\bibitem[Choi et~al.(2018)Choi, Choi, Kim, Ha, Kim, and Choo]{stargan2018}
Yunjey Choi, Minje Choi, Munyoung Kim, Jung-Woo Ha, Sunghun Kim, and Jaegul Choo.
\newblock {StarGAN: Unified Generative Adversarial Networks for Multi-Domain Image-to-Image Translation}.
\newblock In \emph{Proceedings of the IEEE Conference on Computer Vision and Pattern Recognition (CVPR)}, 2018.

\bibitem[Draelos and Carin(2020)]{hirescam}
Rachel~Lea Draelos and Lawrence Carin.
\newblock Use hirescam instead of grad-cam for faithful explanations of convolutional neural networks.
\newblock \emph{arXiv preprint arXiv:2011.08891}, 2020.

\bibitem[Fu et~al.(2020)Fu, Hu, Dong, Guo, Gao, and Li]{xgradcam}
Ruigang Fu, Qingyong Hu, Xiaohu Dong, Yulan Guo, Yinghui Gao, and Biao Li.
\newblock Axiom-based grad-cam: Towards accurate visualization and explanation of cnns.
\newblock \emph{arXiv preprint arXiv:2008.02312}, 2020.

\bibitem[Galbally et~al.(2012)Galbally, Ortiz-Lopez, Fierrez, and Ortega-Garcia]{real_print12}
Javier Galbally, Jaime Ortiz-Lopez, Julian Fierrez, and Javier Ortega-Garcia.
\newblock Iris liveness detection based on quality related features.
\newblock In \emph{2012 5th IAPR International Conference on Biometrics (ICB)}, pages 271--276. IEEE, 2012.

\bibitem[Gao et~al.(2019)Gao, Pei, and Huang]{progan2019}
Hongchang Gao, Jian Pei, and Heng Huang.
\newblock {ProGAN: Network Embedding via Proximity Generative Adversarial Network}.
\newblock In \emph{Proceedings of the 25th ACM SIGKDD International Conference on Knowledge Discovery \& Data Mining}, page 1308–1316, New York, NY, USA, 2019. Association for Computing Machinery.

\bibitem[Heo et~al.(2019)Heo, Joo, and Moon]{heo2019fooling}
Juyeon Heo, Sunghwan Joo, and Taesup Moon.
\newblock Fooling neural network interpretations via adversarial model manipulation.
\newblock \emph{Advances in neural information processing systems}, 32, 2019.

\bibitem[Huang et~al.(2017)Huang, Liu, van~der Maaten, and Weinberger]{densenet121_2017}
Gao Huang, Zhuang Liu, Laurens van~der Maaten, and Kilian~Q. Weinberger.
\newblock Densely connected convolutional networks.
\newblock In \emph{Proceedings of the IEEE Conference on Computer Vision and Pattern Recognition (CVPR)}, 2017.

\bibitem[Johnson et~al.(2019)Johnson, Pollard, Berkowitz, Greenbaum, Lungren, Deng, Mark, and Horng]{mimicxr}
Alistair~EW Johnson, Tom~J Pollard, Seth~J Berkowitz, Nathaniel~R Greenbaum, Matthew~P Lungren, Chih-ying Deng, Roger~G Mark, and Steven Horng.
\newblock Mimic-cxr, a de-identified publicly available database of chest radiographs with free-text reports.
\newblock \emph{Scientific data}, 6\penalty0 (1):\penalty0 317, 2019.

\bibitem[Karras et~al.(2017)Karras, Aila, Laine, and Lehtinen]{celeba_hq2017}
Tero Karras, Timo Aila, Samuli Laine, and Jaakko Lehtinen.
\newblock Progressive growing of gans for improved quality, stability, and variation, 2017.

\bibitem[Karras et~al.(2019{\natexlab{a}})Karras, Laine, and Aila]{FFHQ2018}
Tero Karras, Samuli Laine, and Timo Aila.
\newblock A style-based generator architecture for generative adversarial networks.
\newblock In \emph{Proceedings of the IEEE/CVF conference on computer vision and pattern recognition}, pages 4401--4410, 2019{\natexlab{a}}.

\bibitem[Karras et~al.(2019{\natexlab{b}})Karras, Laine, and Aila]{stylegan2019}
Tero Karras, Samuli Laine, and Timo Aila.
\newblock A style-based generator architecture for generative adversarial networks.
\newblock In \emph{Proceedings of the IEEE/CVF Conference on Computer Vision and Pattern Recognition (CVPR)}, 2019{\natexlab{b}}.

\bibitem[Karras et~al.(2020{\natexlab{a}})Karras, Aittala, Hellsten, Laine, Lehtinen, and Aila]{stylegan_ada2020}
Tero Karras, Miika Aittala, Janne Hellsten, Samuli Laine, Jaakko Lehtinen, and Timo Aila.
\newblock Training generative adversarial networks with limited data.
\newblock In \emph{Proc. NeurIPS}, 2020{\natexlab{a}}.

\bibitem[Karras et~al.(2020{\natexlab{b}})Karras, Laine, Aittala, Hellsten, Lehtinen, and Aila]{stylegan2020}
Tero Karras, Samuli Laine, Miika Aittala, Janne Hellsten, Jaakko Lehtinen, and Timo Aila.
\newblock Analyzing and improving the image quality of stylegan.
\newblock In \emph{Proceedings of the IEEE/CVF conference on computer vision and pattern recognition}, pages 8110--8119, 2020{\natexlab{b}}.

\bibitem[Karras et~al.(2021)Karras, Aittala, Laine, H\"ark\"onen, Hellsten, Lehtinen, and Aila]{stylegan2021}
Tero Karras, Miika Aittala, Samuli Laine, Erik H\"ark\"onen, Janne Hellsten, Jaakko Lehtinen, and Timo Aila.
\newblock Alias-free generative adversarial networks.
\newblock In \emph{Proc. NeurIPS}, 2021.

\bibitem[Kohli et~al.(2013)Kohli, Yadav, Vatsa, and Singh]{real_textured20}
Naman Kohli, Daksha Yadav, Mayank Vatsa, and Richa Singh.
\newblock Revisiting iris recognition with color cosmetic contact lenses.
\newblock In \emph{2013 International Conference on Biometrics (ICB)}, pages 1--7. IEEE, 2013.

\bibitem[Kohli et~al.(2016)Kohli, Yadav, Vatsa, Singh, and Noore]{print21}
Naman Kohli, Daksha Yadav, Mayank Vatsa, Richa Singh, and Afzel Noore.
\newblock Detecting medley of iris spoofing attacks using desist.
\newblock In \emph{2016 IEEE 8th International Conference on Biometrics Theory, Applications and Systems (BTAS)}, pages 1--6. IEEE, 2016.

\bibitem[Lee et~al.(2007)Lee, Park, Lee, Bae, and Kim]{real_artificial_textured_print_24}
Sung~Joo Lee, Kang~Ryoung Park, Youn~Joo Lee, Kwanghyuk Bae, and Jaihie Kim.
\newblock Multifeature-based fake iris detection method.
\newblock \emph{Optical Engineering}, 46\penalty0 (12):\penalty0 127204--127204, 2007.

\bibitem[Muhammad and Yeasin(2020)]{eigencam}
Mohammed~Bany Muhammad and Mohammed Yeasin.
\newblock Eigen-cam: Class activation map using principal components.
\newblock In \emph{2020 international joint conference on neural networks (IJCNN)}, pages 1--7. IEEE, 2020.

\bibitem[of~Technology(2013)]{real43}
Warsaw~University of Technology.
\newblock Warsaw datasets webpage.
\newblock \url{http://zbum.ia.pw.edu.pl/EN/node/46}, 2013.

\bibitem[Phillips et~al.(2017)Phillips, Flynn, and Bowyer]{FRGC}
P~Jonathon Phillips, Patrick~J Flynn, and Kevin~W Bowyer.
\newblock Lessons from collecting a million biometric samples.
\newblock \emph{Image and Vision Computing}, 58:\penalty0 96--107, 2017.

\bibitem[Ramaswamy et~al.(2020)]{ablationcam}
Harish~Guruprasad Ramaswamy et~al.
\newblock Ablation-cam: Visual explanations for deep convolutional network via gradient-free localization.
\newblock In \emph{proceedings of the IEEE/CVF winter conference on applications of computer vision}, pages 983--991, 2020.

\bibitem[Rigas and Komogortsev(2015)]{real22}
Ioannis Rigas and Oleg~V Komogortsev.
\newblock Eye movement-driven defense against iris print-attacks.
\newblock \emph{Pattern Recognition Letters}, 68:\penalty0 316--326, 2015.

\bibitem[Selvaraju et~al.(2017)Selvaraju, Cogswell, Das, Vedantam, Parikh, and Batra]{gradcam}
Ramprasaath~R Selvaraju, Michael Cogswell, Abhishek Das, Ramakrishna Vedantam, Devi Parikh, and Dhruv Batra.
\newblock Grad-cam: Visual explanations from deep networks via gradient-based localization.
\newblock In \emph{Proceedings of the IEEE international conference on computer vision}, pages 618--626, 2017.

\bibitem[Trokielewicz et~al.(2015{\natexlab{a}})Trokielewicz, Czajka, and Maciejewicz]{diseased38}
Mateusz Trokielewicz, Adam Czajka, and Piotr Maciejewicz.
\newblock Assessment of iris recognition reliability for eyes affected by ocular pathologies.
\newblock In \emph{2015 IEEE 7th International Conference on Biometrics Theory, Applications and Systems (BTAS)}, pages 1--6. IEEE, 2015{\natexlab{a}}.

\bibitem[Trokielewicz et~al.(2015{\natexlab{b}})Trokielewicz, Czajka, and Maciejewicz]{real_diseased}
Mateusz Trokielewicz, Adam Czajka, and Piotr Maciejewicz.
\newblock Assessment of iris recognition reliability for eyes affected by ocular pathologies.
\newblock In \emph{2015 IEEE 7th International Conference on Biometrics Theory, Applications and Systems (BTAS)}, pages 1--6. IEEE, 2015{\natexlab{b}}.

\bibitem[Trokielewicz et~al.(2020)Trokielewicz, Czajka, and Maciejewicz]{post40}
Mateusz Trokielewicz, Adam Czajka, and Piotr Maciejewicz.
\newblock Post-mortem iris recognition with deep-learning-based image segmentation.
\newblock \emph{Image and Vision Computing}, 94:\penalty0 103866, 2020.

\bibitem[Wang et~al.(2020)Wang, Wang, Du, Yang, Zhang, Ding, Mardziel, and Hu]{scorecam}
Haofan Wang, Zifan Wang, Mengnan Du, Fan Yang, Zijian Zhang, Sirui Ding, Piotr Mardziel, and Xia Hu.
\newblock Score-cam: Score-weighted visual explanations for convolutional neural networks.
\newblock In \emph{Proceedings of the IEEE/CVF conference on computer vision and pattern recognition workshops}, pages 24--25, 2020.

\bibitem[Wei et~al.(2008)Wei, Tan, and Sun]{synth44}
Zhuoshi Wei, Tieniu Tan, and Zhenan Sun.
\newblock Synthesis of large realistic iris databases using patch-based sampling.
\newblock In \emph{2008 19th International Conference on Pattern Recognition}, pages 1--4. IEEE, 2008.

\bibitem[Yambay et~al.({\natexlab{a}})Yambay, Becker, Kohli, Yadav, Czajka, Bowyer, Schuckers, Singh, Vatsa, Noore, et~al.]{real_textured45}
David Yambay, Benedict Becker, Naman Kohli, Daksha Yadav, Adam Czajka, Kevin~W Bowyer, Stephanie Schuckers, Richa Singh, Mayank Vatsa, Afzel Noore, et~al.
\newblock Livdet iris 2017-iris liveness detection competition 2017.
\newblock {\natexlab{a}}.

\bibitem[Yambay et~al.({\natexlab{b}})Yambay, Walczak, Schuckers, and Czajka]{real_textured46}
David Yambay, Brian Walczak, Stephanie Schuckers, and Adam Czajka.
\newblock Livdet-iris 2015--iris liveness detection competition 2015.
\newblock {\natexlab{b}}.

\bibitem[Zhou et~al.(2016{\natexlab{a}})Zhou, Khosla, Lapedriza, Oliva, and Torralba]{Zhou_CVPR_2016}
Bolei Zhou, Aditya Khosla, Agata Lapedriza, Aude Oliva, and Antonio Torralba.
\newblock Learning deep features for discriminative localization.
\newblock In \emph{2016 IEEE Conference on Computer Vision and Pattern Recognition (CVPR)}, pages 2921--2929, 2016{\natexlab{a}}.

\bibitem[Zhou et~al.(2016{\natexlab{b}})Zhou, Khosla, Lapedriza, Oliva, and Torralba]{zhou2016initialCAM}
Bolei Zhou, Aditya Khosla, Agata Lapedriza, Aude Oliva, and Antonio Torralba.
\newblock Learning deep features for discriminative localization.
\newblock In \emph{Proceedings of the IEEE conference on computer vision and pattern recognition}, pages 2921--2929, 2016{\natexlab{b}}.

\end{thebibliography}
